\documentclass[runningheads]{llncs}
\usepackage[T1]{fontenc}
\usepackage{graphicx}
\usepackage{amsmath,amssymb}
\usepackage{multirow}
\usepackage{xcolor}
\usepackage[breaklinks,colorlinks,allcolors=blue]{hyperref}
\usepackage[capitalize,noabbrev]{cleveref}
\begin{document}
%
% \title{FetSelect: Task-Specific Architectures and Self-Supervised Learning for Automated Fetal Ultrasound Frame Selection}
% \titlerunning{FetSelect: Automated Fetal Ultrasound Frame Selection}
% %
% \author{Mahmood Alzubaidi\inst{1} \and Raden Muaz\inst{1} \and Uzair Shah\inst{1}
% \and Mohammed Ammar\inst{2} \and Khalid Alyafei\inst{3}
% \and Mowafa Househ\inst{1} \and Marco Agus\inst{1}}
% \authorrunning{M. Alzubaidi et al.}
% \institute{College of Science and Engineering, Hamad Bin Khalifa University, Doha, Qatar\\
% \email{\{malzubaidi,rara89608,uzsh31989,mhouseh,magus\}@hbku.edu.qa}
% \and LIST Laboratory Department of Electrical Systems Engineering, University of Boumerdes, Boumerdes, Algeria\\
% \email{ammar.mohammed4@gmail.com}
% \and Sidra Medicine, Doha, Qatar\\
% \email{kalyafei@sidra.org}}

\title{FetSelect: Task-Specific Architectures and Self-Supervised Learning for Automated Fetal Ultrasound Frame Selection}
\titlerunning{FetSelect: Automated Fetal Ultrasound Frame Selection}
\author{Mahmood Alzubaidi\inst{1} \and Raden Muaz\inst{1} \and Uzair Shah\inst{1}
\and Mohammed Ammar\inst{2} \and Khalid Alyafei\inst{3}
\and Mowafa Househ\inst{1} \and Marco Agus\inst{1}}
\authorrunning{M. Alzubaidi et al.}
\institute{College of Science and Engineering, Hamad Bin Khalifa University, Doha, Qatar\\
\email{}
\and LIST Laboratory Department of Electrical Systems Engineering, University of Boumerdes, Boumerdes, Algeria\\
\email{}
\and Sidra Medicine, Doha, Qatar\\
\email{}}

\maketitle
\begin{abstract}
Automated frame selection for fetal biometry remains under-addressed, with most prior work targeting generic quality assessment or downstream measurement pipelines that assume suitable frames are available. We introduce \textit{FetSelect}, a task-specific framework that pairs a frozen vision foundation backbone with a hybrid multi-head design: a Task-Gated classification head and a Detection-derived quality head combined via learned fusion. We curate 6{,}486 expert-labeled frames across four targets---Crown--Rump Length (CRL), Nuchal Translucency (NT), Nasal Bone (NB), and Scalebar---and adapt the backbone with BYOL pretraining on 19{,}019 unlabeled images. On a held-out test set (974 frames), FetSelect achieves mean AUROC \textbf{0.956} and mean correlation \textbf{0.818} with expert quality annotations. Ablations confirm that hybrid fusion surpasses single-head variants, and ultrasound-specific self-supervision yields consistent gains. Evaluation on external clinical videos and 509 external CRL images demonstrates task-specific discrimination.

\end{abstract}

%% ============================================================================
\section{Introduction}
\label{sec:intro}
%% ============================================================================

Selecting diagnostically optimal frames from fetal ultrasound videos is a prerequisite for reliable first-trimester screening. In routine practice, sonographers sweep through the midsagittal view and manually pick frames for crown--rump length (CRL), nuchal translucency (NT), nasal bone (NB) assessment, and scalebar calibration~\cite{9761400,persico2012nasal}. Missed or suboptimal frames degrade measurements, increase operator burden, and complicate quality control~\cite{Plotka2025,Zhang2021AutoFSQuality,venturini2024whole}. Although standard-plane detection~\cite{baumgartner2017sononet} and end-to-end biometry pipelines~\cite{venturini2024whole,ramesh2025automated} have advanced markedly, most assume expert pre-selection of frames or optimize a single downstream endpoint rather than the \emph{upstream} selection of task-specific, diagnostically optimal frames. Related video-to-measurement works~\cite{PLOTKA2023101182} and first-trimester CRL/NT algorithms~\cite{9761400} still depend on having an appropriate frame available.

\begin{figure}[t]
  \centering
  \includegraphics[width=0.95\textwidth]{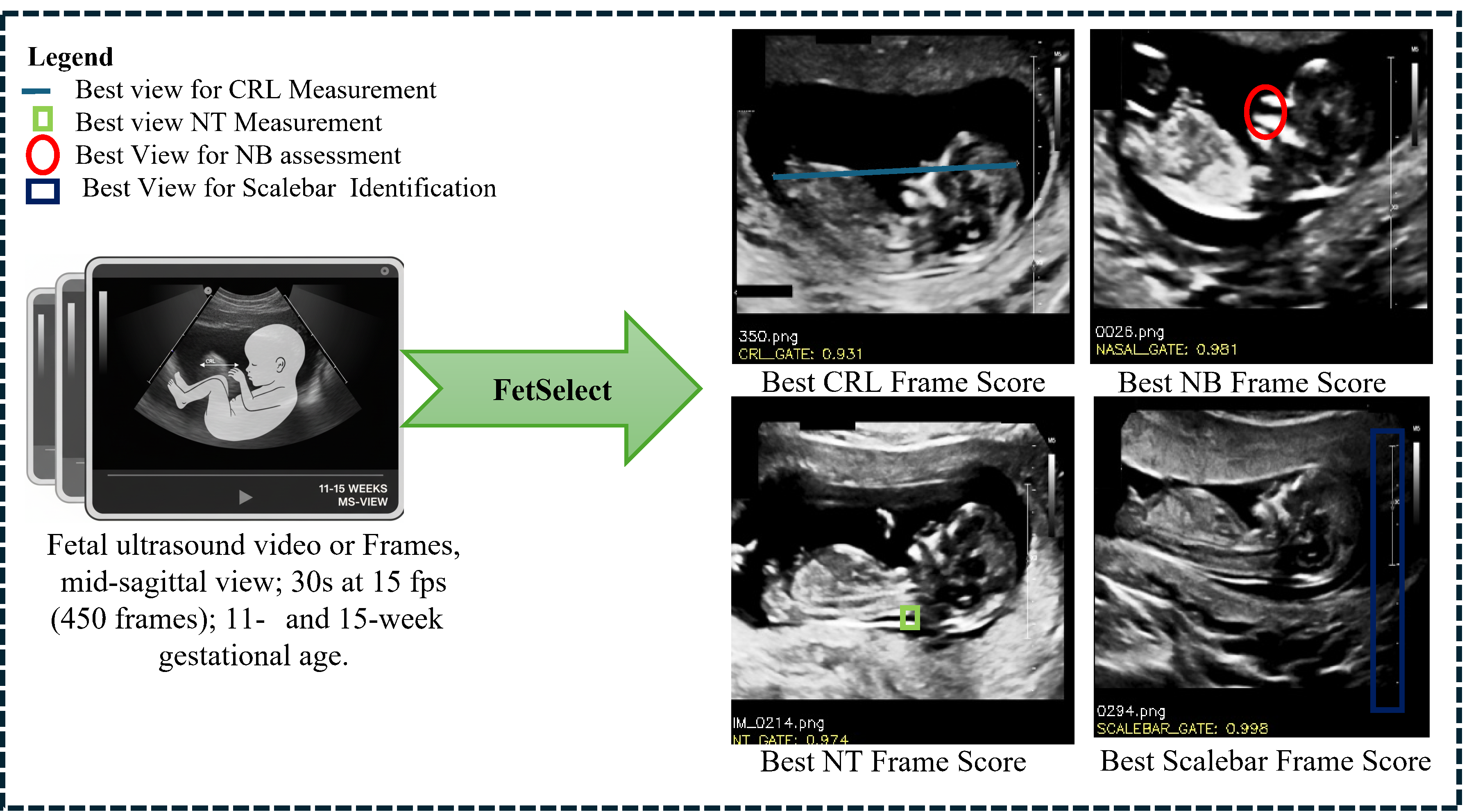}
  \caption{\textbf{FetSelect overview.} FetSelect ingests midsagittal ultrasound videos or frames and outputs task-specific frame-quality scores for CRL, NT, NB, and scalebar; the highest-scoring frames are selected for clinical use.}
  \label{fig:intro_teaser}
\end{figure}

Medical image quality assessment~\cite{deep2020}, ultrasound frame selection~\cite{xu2022adaptive,guo2024mmsummary,singh2022metric}, clip retrieval~\cite{MISHRA2025103611}, and foundation-model IQA~\cite{he2025advancingfetalultrasoundimage} typically emphasize global quality or single endpoints and do not enforce \emph{clinical criteria that differ across CRL, NT, NB, and scalebar}. Selection criteria diverge across tasks: CRL requires a full fetal profile in neutral posture; NT demands precise midsagittal delineation of a 1--3\,mm hypoechoic space; NB is sensitive to the exact midsagittal plane~\cite{persico2012nasal}; scalebar detection depends on stable placement and legible tick marks.

\paragraph{Our approach.}
We introduce \emph{FetSelect} (\cref{fig:intro_teaser}), a unified framework for automated \emph{task-specific} frame selection across four first-trimester objectives. FetSelect combines (i)~\emph{ultrasound-specific} self-supervision with BYOL on 19{,}019 unlabeled images, and (ii)~a lightweight multi-path head trained on 6{,}486 multi-annotated frames. The head fuses a \emph{task-gated} pathway (global features) with a \emph{detection-derived} pathway (localization statistics), and a learned \emph{fusion} module; the frozen backbone and small heads ($\sim$2.2M params) enable real-time inference.

\paragraph{Contributions.}
\begin{itemize}
  \item We cast fetal ultrasound \emph{frame selection} as a multi-objective, task-specific learning problem spanning CRL, NT, NB, and scalebar---the upstream bottleneck preceding biometry.
  \item A new multi-task dataset of \textbf{6,486} first-trimester frames with multi-modal annotations (boxes, keypoints, polygons, clinical attributes).
  \item A multi-path head combining a task-gated classifier, a detection-derived quality converter, and a learned fusion module (\cref{sec:methods}).
\end{itemize}

%% ============================================================================
\section{Related Work}
\label{sec:related}
%% ============================================================================

\paragraph{Fetal ultrasound analysis and biometry.}
SonoNet~\cite{baumgartner2017sononet} introduced real-time standard-plane recognition with weakly supervised localization, establishing the paradigm of end-to-end view classification from video frames. Subsequent pipelines process entire examinations to aggregate noisy measurements into calibrated biometrics~\cite{venturini2024whole}, while related workflows automate intrapartum measurements across classification, segmentation, and measurement stages~\cite{ramesh2025automated}. Multi-task fetal biometry from ultrasound videos~\cite{PLOTKA2023101182} demonstrates that direct video-to-measurement paradigms are feasible, and task-specific first-trimester efforts include standard-plane localization for CRL and NT~\cite{9761400}. Nasal bone assessment is particularly sensitive to the exact midsagittal plane~\cite{persico2012nasal}, underscoring the need for task-specific criteria when selecting frames. However, these pipelines assume expert pre-selection of suitable frames or optimize only downstream measurements rather than the upstream selection of diagnostically optimal frames.

\paragraph{Frame selection and image quality assessment.}
Medical image quality assessment has a rich literature spanning generic assessment~\cite{deep2020} and domain-specific applications. In ultrasound, frame selection has been explored for obstetric sweeps using adaptive temporal selection~\cite{xu2022adaptive}, multimodal clinical summarization~\cite{guo2024mmsummary}, and metric-based selection in lung ultrasound videos~\cite{singh2022metric}. Clip-level retrieval adapts spatiotemporal transformers and contrastive objectives to robustly retrieve standard-frame clips under noise and boundary uncertainty~\cite{MISHRA2025103611}. Foundation-model adaptation has been leveraged for IQA on blind sweeps~\cite{he2025advancingfetalultrasoundimage}. However, these methods emphasize global quality or single endpoints and seldom enforce \emph{task-specific, clinically grounded selection criteria} that differ across CRL, NT, NB, and scalebar.

\paragraph{Vision foundation models and self-supervised learning.}
Vision foundation models show strong transfer to medical imaging. C-RADIO-B~\cite{cradio2024}---a Vision Transformer distilled from CLIP, DINOv2, and SAM---provides rich multi-scale features suitable for downstream medical tasks. General-purpose self-supervised models such as DINOv2 show mixed, modality-dependent outcomes in medical domains~\cite{oquab2023dinov2}, motivating domain-specific adaptation. Fetal-specific pretraining is emerging: FetalCLIP~\cite{maani2025fetalclip}, trained on 210K image--text pairs, yields robust downstream performance across tasks with limited labels, and parameter-efficient adaptations further benefit blind-sweep IQA~\cite{he2025advancingfetalultrasoundimage}. Concurrent ultrasound-specific foundation models such as USFM~\cite{jiao2024usfm} and USF-MAE~\cite{usfmae2024} pretrain on large unlabeled ultrasound corpora across organs, but target generic representation learning rather than the task-specific frame-selection problem we address. Video-based ultrasound SSL demonstrates that temporally aware objectives improve downstream tasks~\cite{jiao2020self,stebler2025temporalrepresentationlearningrealtime}. We compare four canonical SSL objectives---BYOL~\cite{bootstrap2020}, MoCo~\cite{momentum2020}, DINO~\cite{Dino_loss}, and VICReg~\cite{bardes2022vicreg}---finding that ultrasound-specific BYOL pretraining yields the most consistent gains for frame selection, while the remaining objectives provide informative comparisons for augmentation sensitivity and architectural compatibility.

\paragraph{Summary and gap.}
Existing work optimizes downstream biometry or global IQA and typically presumes optimal frames are available. Foundation models and generic SSL alone are insufficient for specialized ultrasound selection without domain adaptation. Multi-task learning with modality-aware fusion and task-specific architectures has been explored broadly in medical imaging, but a systematic comparison of fusion designs for \emph{frame selection} in fetal ultrasound is lacking. \emph{FetSelect} fills this gap with a unified, task-specific framework that combines ultrasound-specific SSL with a lightweight fusion head integrating global context and detection-derived anatomical evidence.

%% ============================================================================
\section{Methods and Materials}
\label{sec:methods}
%% ============================================================================

\subsection{Datasets}
\label{subsec:datasets}

\paragraph{Self-supervised learning dataset.}
We compiled a pretraining corpus of 19,019 fetal ultrasound images from 14 anatomical categories (abdomen, aorta, CRL view, cervical, cervix, femur, NT view, non-standard/standard NT, pubic symphysis--fetal head, thorax, transcerebellar, transthalamic, transventricular) sourced from public repositories~\cite{da_correggio_2023_fetal,Burgos-ArtizzuFetalPlanesDataset,ALZUBAIDI2023109708,LIN2022107170,Chen2024} and Kaggle. The corpus spans imaging protocols, gestational ages, and clinical contexts. All SSL images use the same preprocessing as the supervised set: resize to $224\times 224$, per-image intensity normalization, and cropping of vendor banner overlays; only quality labels are added in the supervised pipeline. We deliberately span all three trimesters in the SSL corpus---even though downstream tasks are first-trimester only---because the wider anatomical and probe-pose distribution gives the frozen backbone richer ultrasound-specific texture and structural priors. Domain shift between the SSL and supervised sets is therefore limited to anatomical class distribution rather than acquisition or preprocessing.

\paragraph{Supervised learning dataset.}
For supervised training and evaluation, we curated a multi-task annotated dataset of 6,486 ultrasound frames across four first-trimester tasks: CRL, NT, NB, and scalebar calibration (\cref{tab:supervised_dataset}). The corpus includes bounding boxes, keypoints, polygons, and clinical attributes, partitioned into training (70\%), validation (15\%), and test (15\%) splits via stratified sampling.

\begin{table}[t]
\caption{Supervised learning dataset statistics. The annotated corpus encompasses
6{,}486 frames with multi-modal annotations across four clinical tasks.}\label{tab:supervised_dataset}
\centering
\begin{tabular}{|l|c|c|c|c|}
\hline
\textbf{Task} & \textbf{Frames} & \textbf{Train} & \textbf{Val} & \textbf{Test} \\
\hline
NT       & 517   & 362   & 78  & 77  \\
CRL      & 1,169 & 818   & 175 & 176 \\
NB       & 4,239 & 2,967 & 636 & 636 \\
Scalebar & 561   & 393   & 84  & 84  \\
\hline
\textbf{Total} & \textbf{6,486} & \textbf{4,540} & \textbf{972} & \textbf{974} \\
\hline
\end{tabular}
\end{table}

\paragraph{Annotation quality and label definition.}
The dataset was annotated by a senior sonographer (4 years experience). We employ rule-guided scoring (\cref{eq:quality_heuristic}) to produce continuous quality labels. We acknowledge that these heuristics are a proxy for real clinical quality; however, they incorporate domain-specific criteria mirroring established clinical scoring systems~\cite{Plotka2025} (fluid clarity for NT, neutral-posture full profile for CRL, bone definition for NB, tick legibility for scalebar). The detection pathway (Path~B) uses \textit{predicted} detections at inference---not ground-truth annotations---reducing label circularity. Inter-observer agreement (50 images/task, two senior experts with 10+ years experience) yielded Cohen's $\kappa$: CRL 0.82, NT 0.76, NB 0.79, Scalebar 0.92. Intra-observer ICC was 0.88.

\subsection{FetSelect Architecture}
\label{subsec:architecture}

\begin{figure}[t]
\centering
\includegraphics[width=\textwidth]{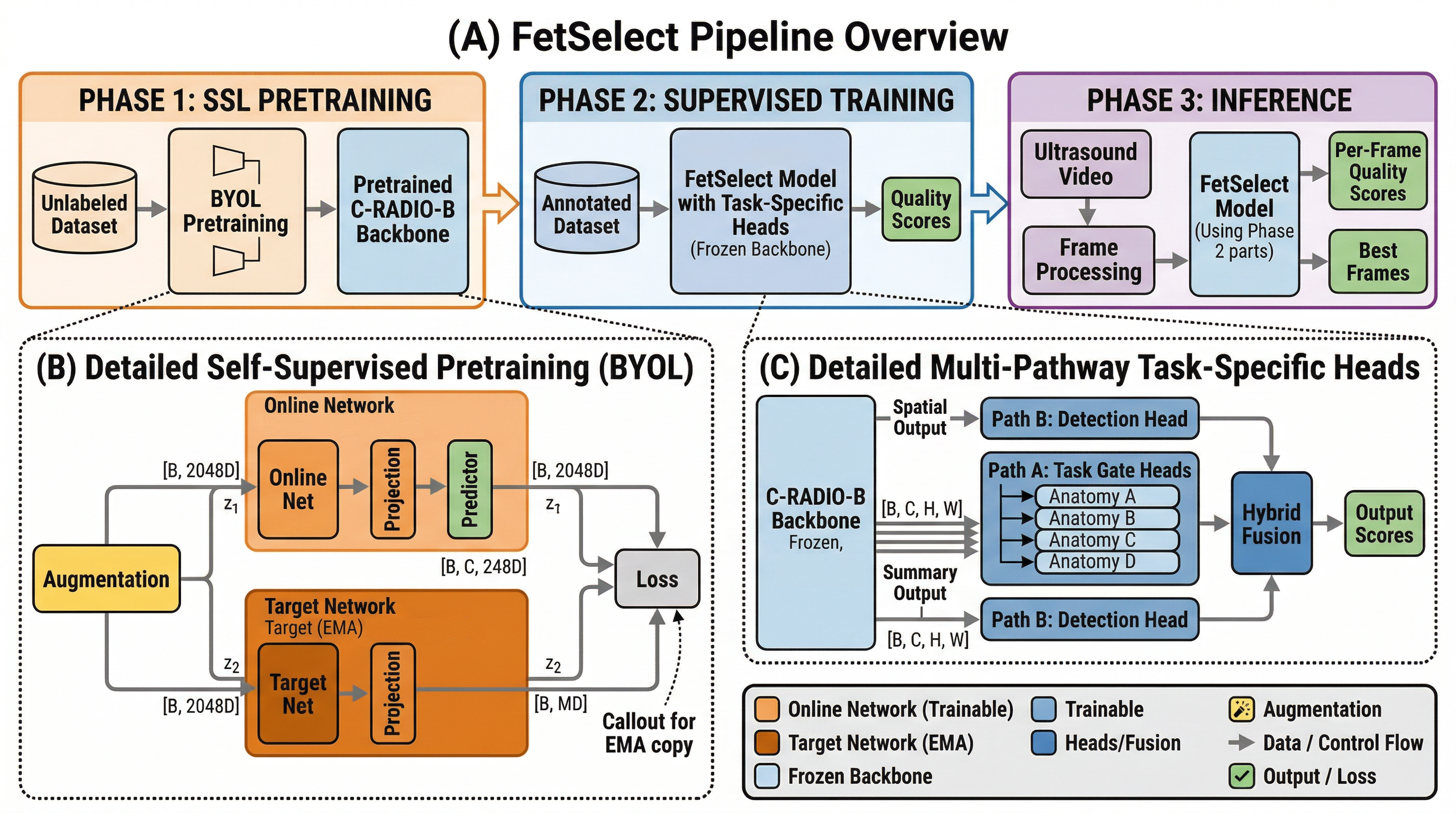}
\caption{\textbf{FetSelect architecture.} \textbf{Phase~1}: BYOL pretraining on 19K unlabeled images with C-RADIO-B backbone. \textbf{Phase~2}: Supervised training of Task-Gated, Detection, and Fusion heads (backbone frozen). \textbf{Phase~3}: Inference-time frame ranking across four clinical tasks.}
\label{fig:fetselect_architecture}
\end{figure}

FetSelect comprises three phases (\cref{fig:fetselect_architecture}): (1)~self-supervised backbone pretraining, (2)~supervised training of task-specific heads, and (3)~inference-time frame ranking.

\paragraph{Self-supervised pretraining with BYOL.}
We employed Bootstrap Your Own Latent (BYOL)~\cite{bootstrap2020}, a non-contrastive framework avoiding collapse without negative pairs. BYOL uses an \textit{online network} ($\theta$) and a \textit{target network} ($\xi$) sharing identical encoder architectures. Both utilize C-RADIO-B~\cite{cradio2024,Heinrich_2025_CVPR}, a 98.5M-parameter Vision Transformer distilled from CLIP, DINOv2, and SAM, producing 2304-dimensional features. Crucially, the online network includes a \emph{predictor head} $q_\theta$ absent from the target network---this asymmetry is the mechanism preventing representation collapse~\cite{bootstrap2020}.

Given an image $x$, two augmented views $v_1, v_2$ are generated using random flipping, rotation ($\pm 10^\circ$), brightness/contrast adjustment ($\pm 20\%$), multiplicative Rayleigh noise, and elastic deformations. We model ultrasound speckle as multiplicative Rayleigh noise ($I' = I \cdot n$, $n \sim \text{Rayleigh}(\sigma)$) rather than additive Gaussian noise, since speckle arises from constructive/destructive interference of backscattered echoes and is inherently multiplicative.

The online network computes projection $z_\theta^i = g_\theta(f_\theta(v_i))$ and prediction $p_\theta^i = q_\theta(z_\theta^i)$; the target computes only $z_\xi^i = g_\xi(f_\xi(v_i))$. The symmetrized BYOL loss is:
\begin{equation}
\mathcal{L}_{\text{BYOL}} = \left\| \frac{q_\theta(z_\theta^1)}{\|q_\theta(z_\theta^1)\|_2} - \frac{z_\xi^2}{\|z_\xi^2\|_2} \right\|_2^2 + \left\| \frac{q_\theta(z_\theta^2)}{\|q_\theta(z_\theta^2)\|_2} - \frac{z_\xi^1}{\|z_\xi^1\|_2} \right\|_2^2
\label{eq:byol_loss}
\end{equation}
where the predictor $q_\theta$ appears explicitly, applied only to the online branch. Target parameters are updated via EMA: $\xi \leftarrow \tau \xi + (1 - \tau) \theta$ with $\tau = 0.996$. We pretrained on 19,019 images for 100 epochs (batch size 12, LR $10^{-3}$, cosine decay, AdamW, 10-epoch warmup; converged at epoch 40).

\paragraph{Path A: Task-Gated Head.}
Given frozen backbone outputs---summary features $\mathbf{s} \in \mathbb{R}^{2304}$ and spatial features $\mathbf{F}_{\text{spatial}} \in \mathbb{R}^{768 \times 16 \times 16}$---we pool spatial features and concatenate ($\mathbf{h} \in \mathbb{R}^{3072}$). Per-task two-layer MLPs (ReLU, dropout $p{=}0.2$) output quality logits $y^{(k)}_{\text{gate}}$. Ground-truth labels are computed via heuristic quality scoring:
\begin{equation}
q^{(k)} = w_1^{(k)} \cdot \mathbb{1}_{\text{present}} + w_2^{(k)} \cdot c^{(k)} + w_3^{(k)} \cdot v^{(k)}
\label{eq:quality_heuristic}
\end{equation}
where $\mathbb{1}_{\text{present}}$ is binary annotation presence, $c^{(k)} \in [0,1]$ is the fraction of required landmarks visible, and $v^{(k)} \in [0,1]$ is a task-specific visibility ratio computed deterministically from box and keypoint geometry (e.g., NT-segment width relative to expected range; tick legibility for scalebar). All three terms are fixed inputs derived from annotations and are not learnable. The per-task weights $w_1^{(k)}, w_2^{(k)}, w_3^{(k)}$ are clinical hyperparameters set prior to training; only the task-gated MLP weights are optimized. Final scores are normalized to $[0,1]$.

\paragraph{Path B: Detection Head with Quality Converter.}
\label{par:path_b}
Anchor-based detection on spatial features produces bounding boxes and class logits across 3 scales. A detection-to-gate converter extracts five statistics per task (max confidence, count, average/max area, coverage) and maps them to quality logits $y^{(k)}_{\text{det}}$ via task-specific networks. This pathway uses \emph{predicted} detections at inference, not ground-truth annotations.

\paragraph{Fusion module and loss.}
Per-task fusion networks combine $[y^{(k)}_{\text{gate}}; y^{(k)}_{\text{det}}]$ to produce fused logits $y^{(k)}_{\text{fused}}$. All pathways are trained jointly:
\begin{equation}
\mathcal{L}_{\text{total}} = \sum_{k=1}^{4} \Big[ \mathcal{L}_{\text{BCE}}(y^{(k)}_{\text{gate}}, \hat{y}^{(k)}) + \mathcal{L}_{\text{BCE}}(y^{(k)}_{\text{det}}, \hat{y}^{(k)}) + \mathcal{L}_{\text{BCE}}(y^{(k)}_{\text{fused}}, \hat{y}^{(k)}) \Big]
\end{equation}
where $\mathcal{L}_{\text{BCE}}$ is binary cross-entropy with logits and $\hat{y}^{(k)} \in [0, 1]$ are ground-truth scores. BCE was selected over focal loss after controlled comparison: focal loss yielded AUROC 0.870 vs.\ BCE 0.953 on C-RADIO-B with fusion, likely because focal loss down-weights easy examples that still carry useful gradient signal for continuous quality regression. Training used 4,540 frames, 100 epochs with early stopping (patience 10), batch size 12 with 4-step accumulation, LR $10^{-4}$, cosine annealing, AdamW. The frozen backbone (98.5M) remained fixed; task heads (2.2M) were optimized.

\paragraph{Inference efficiency.}
With the backbone frozen and only the lightweight heads active ($\sim$2.2M parameters), single-frame inference runs at $\sim$111\,FPS on an NVIDIA RTX 4090 (latency $9.0 \pm 0.0$\,ms, batch size 1). This exceeds typical clinical ultrasound frame rates (15--30\,FPS) by $>$3$\times$, enabling real-time frame scoring during live acquisition without interrupting the clinical workflow. On CPU (Intel i7-12700, single thread, ONNX runtime), the heads alone run at $\sim$50\,FPS once backbone features are cached; the full pipeline including the frozen C-RADIO-B backbone runs at $\sim$1.8\,FPS, which is suited to offline batch scoring of stored sweeps and to retrospective quality control rather than live acquisition. A distilled or quantized backbone is the natural next step for live CPU deployment.

\subsection{Evaluation Metrics}
\label{subsec:metrics}

Per-task and mean: (1)~\textbf{AUROC} (binarized at 0.5); (2)~\textbf{Average Precision (AP)}; (3)~\textbf{Pearson Correlation}. The \textbf{composite score} $= \tfrac{1}{2}(\overline{\text{AUROC}} + \overline{\text{Correlation}})$.

%% ============================================================================
\section{Results and Evaluation}
\label{sec:results}
%% ============================================================================

\subsection{Performance on Supervised Test Set}
\label{subsec:main_results}

\Cref{tab:main_results} reports performance on the held-out test set (974 frames). FetSelect attains mean AUROC \textbf{0.956}, mean AP \textbf{0.789}, mean Corr.\ \textbf{0.818}, and composite \textbf{0.887}. CRL (0.996) and NB (0.988) show near-perfect discrimination; NT (0.931) and Scalebar (0.908) are more challenging but still excellent.

\begin{table}[t]
\caption{FetSelect performance on the held-out test set (974 frames). All values
report mean $\pm$ std over 3 random seeds.}\label{tab:main_results}
\centering
\begin{tabular}{|l|c|c|c|}
\hline
\textbf{Task} & \textbf{AUROC} & \textbf{AP} & \textbf{Corr.} \\
\hline
NT         & $0.931 \pm 0.003$ & $0.286 \pm 0.008$ & $0.745 \pm 0.011$ \\
CRL        & $0.996 \pm 0.001$ & $0.935 \pm 0.004$ & $0.913 \pm 0.006$ \\
Nasal Bone & $0.988 \pm 0.001$ & $0.952 \pm 0.003$ & $0.886 \pm 0.007$ \\
Scalebar   & $0.908 \pm 0.004$ & $0.426 \pm 0.010$ & $0.728 \pm 0.009$ \\
\hline
\textbf{Mean} & $\mathbf{0.956 \pm 0.002}$ & $\mathbf{0.789 \pm 0.006}$ & $\mathbf{0.818 \pm 0.008}$ \\
\hline
\end{tabular}
\end{table}

\subsection{Qualitative Validation on External Videos}
\label{subsec:video_validation}

We assess temporal behavior on four external clips (Samsung HS40, 1232$\times$924, 30 FPS, $\sim$62 frames each). \Cref{tab:qualitative_videos} reports selected frame indices and quality scores. CRL-focused videos (a--c) yield high CRL scores (.884--.915) with low NT scores (.045--.082); the NT-focused video (d) assigns the highest score to NT (.823). Different tasks select different frames within each clip, confirming task-specific criteria.

\begin{table}[t]
\caption{Qualitative validation on external videos. For each task, the best frame index and its quality score are reported.}\label{tab:qualitative_videos}
\centering
\begin{tabular}{|l|c|c|c|c|c|c|c|}
\hline
\multirow{2}{*}{\textbf{Video}} & \multirow{2}{*}{\textbf{Total}} & \multicolumn{2}{c|}{\textbf{CRL}} & \multicolumn{2}{c|}{\textbf{NT}} & \multicolumn{2}{c|}{\textbf{NB}} \\
\cline{3-8}
& & \textbf{Frame} & \textbf{Score} & \textbf{Frame} & \textbf{Score} & \textbf{Frame} & \textbf{Score} \\
\hline
(a) CRL-1 & 62 & 5 & 0.902 & 12 & 0.045 & 8 & 0.231 \\
(b) CRL-2 & 62 & 31 & 0.884 & 28 & 0.067 & 29 & 0.189 \\
(c) CRL-3 & 62 & 18 & 0.915 & 22 & 0.082 & 19 & 0.267 \\
(d) NT-1  & 62 & 42 & 0.612 & 38 & 0.823 & 41 & 0.445 \\
\hline
\end{tabular}
\end{table}

\subsection{Generalization to External Public Fetal Images}
\label{subsec:external_images}

On 509 external CRL images~\cite{Ghelichoghli2021}, CRL achieves the highest mean score (0.674) while non-target tasks score low (NT: .107; Scalebar: .064), confirming task specificity (\cref{tab:external_validation}). Only 13.2\% surpass the high-quality threshold ($>0.8$), consistent with heterogeneous public data.

\begin{table}[t]
\caption{Quality scores on 509 external CRL images (unseen during training).}\label{tab:external_validation}
\centering
\begin{tabular}{|l|c|c|c|}
\hline
\textbf{Task} & \textbf{Mean} & \textbf{SD} & \textbf{High-Qual. ($> 0.8$)} \\
\hline
CRL      & 0.674 & 0.142 & 13.2\% \\
NT       & 0.107 & 0.056 & 0.0\% \\
NB       & 0.401 & 0.037 & 0.0\% \\
Scalebar & 0.064 & 0.047 & 0.0\% \\
\hline
\textbf{Overall} & \textbf{0.312} & \textbf{0.260} & \textbf{3.3\%} \\
\hline
\end{tabular}
\end{table}

\subsection{Self-Supervised Pretraining Comparison}
\label{subsec:ablation_ssl}

\Cref{tab:ablation_ssl} compares SSL methods (identical supervised training; composite $= (\overline{\text{AUROC}} + \overline{\text{Corr.}})/2$). DINO achieves the highest single-run composite (0.870), while BYOL attains 0.860$\pm$0.009 across three seeds. BYOL was selected for the final pipeline due to: (1)~verified reproducibility ($\pm$0.008 AUROC over 3 seeds vs.\ single-run estimates for others); (2)~training stability without negative pairs (batch size 12); and (3)~complementarity with fusion---BYOL + fusion yields the best overall result (AUROC 0.956, composite 0.887; \cref{tab:main_results}), a +0.018 gain over baseline + fusion (0.869; \cref{tab:ablation_arch}).

\textbf{Statistical significance.} Paired $t$-tests (BYOL vs.\ baseline, 3 seeds): CRL $p{=}0.031$, NB $p{=}0.042$, NT $p{=}0.078$, Scalebar $p{=}0.091$; aggregate mean AUROC $p{=}0.038$. Effect sizes (Cohen's $d$) are moderate to large ($d = 0.6$--$1.1$). With $n{=}3$ seeds, statistical power is limited; the consistent directionality and effect sizes support practical significance.

\begin{table}[t]
\caption{Self-supervised pretraining comparison. All rows: frozen C-RADIO-B, multi-head fusion, BCE loss. Baseline and BYOL report mean over 3 seeds; others are single runs.}\label{tab:ablation_ssl}
\centering
\begin{tabular}{|l|c|c|c|c|}
\hline
\textbf{Method} & \textbf{AUROC} & \textbf{AP} & \textbf{Corr.} & \textbf{Comp.} \\
\hline
Baseline (no SSL) & 0.936 & 0.757 & 0.770 & 0.853 \\
VICReg~\cite{bardes2022vicreg} & 0.924 & 0.713 & 0.764 & 0.844 \\
MoCo~\cite{momentum2020} & 0.926 & 0.718 & 0.770 & 0.848 \\
DINO~\cite{Dino_loss} & 0.946 & 0.760 & 0.794 & 0.870 \\
\hline
\textbf{BYOL}~\cite{bootstrap2020} & \textbf{0.940} & \textbf{0.776} & \textbf{0.779} & \textbf{0.860} \\
\hline
\end{tabular}
\end{table}

\subsection{Architecture Ablation and Baseline Comparison}
\label{subsec:ablation_arch}

\textbf{SSL vs.\ extended supervised training.} Unfreezing the final transformer block (30 epochs, LR $10^{-5}$) yielded AUROC 0.894 vs.\ frozen baseline 0.953 and BYOL 0.956, confirming SSL provides value beyond supervised fine-tuning.

\Cref{tab:ablation_arch} ablates architectural components (no SSL) and compares against adapted baselines. No existing method directly addresses multi-task frame selection for first-trimester ultrasound (SonoNet~\cite{baumgartner2017sononet} targets second/third-trimester plane classification; FetalCLIP~\cite{maani2025fetalclip} targets general representation), so we construct baselines from our pipeline components.

\begin{table}[t]
\caption{Architecture ablation and baseline comparison (frozen C-RADIO-B). Composite $= (\overline{\text{AUROC}} + \overline{\text{Corr.}})/2$. Baseline+Fusion and FetSelect report mean over 3 seeds.}\label{tab:ablation_arch}
\centering
\begin{tabular}{|l|c|c|c|c|}
\hline
\textbf{Configuration} & \textbf{AUROC} & \textbf{AP} & \textbf{Corr.} & \textbf{Comp.} \\
\hline
Random & 0.500 & 0.249 & 0.001 & 0.251 \\
TaskGate Only (Linear) & 0.899 & 0.680 & 0.712 & 0.806 \\
Detection Only & 0.876 & 0.702 & 0.691 & 0.784 \\
Baseline+Fusion (no SSL) & 0.945 & 0.779 & 0.793 & 0.869 \\
Swapped Inputs & 0.917 & 0.702 & 0.735 & 0.826 \\
\hline
\textbf{FetSelect (BYOL+Fusion)} & \textbf{0.956} & \textbf{0.789} & \textbf{0.818} & \textbf{0.887} \\
\hline
\end{tabular}
\end{table}

Fusion improves over either single pathway (+0.069 AUROC over Detection-Only; +0.046 over TaskGate-Only). Swapping inputs degrades all tasks, confirming proper feature routing. BYOL pretraining adds a further +0.018 composite over baseline+fusion (0.887 vs.\ 0.869), with the largest gains on NT and Scalebar.

\paragraph{Per-task breakdown.}
\Cref{tab:pertask_auroc} disaggregates AUROC by task for architectural variants (no SSL). Detection-Only dominates structure-dependent tasks where localization cues are strong (CRL 0.976, NB 0.990). However, fusion is critical for the two most challenging tasks: NT (+1.2\% over Detection-Only) and Scalebar (+2.2\%), where global context and local detections are complementary. Swapping pathway inputs degrades all four tasks uniformly, confirming that the task-gated and detection pathways encode distinct, non-interchangeable information.

\begin{table}[t]
\caption{Per-task AUROC for architectural variants (test split, no SSL). Fusion provides the largest gains on the two most challenging tasks: NT and Scalebar.}\label{tab:pertask_auroc}
\centering
\begin{tabular}{|l|c|c|c|c|}
\hline
\textbf{Configuration} & \textbf{NT} & \textbf{CRL} & \textbf{NB} & \textbf{Scalebar} \\
\hline
TaskGate Only       & 0.702 & 0.940 & 0.985 & 0.670 \\
Detection Only      & 0.812 & 0.976 & 0.990 & 0.803 \\
\textbf{With Fusion} & \textbf{0.824} & \textbf{0.976} & \textbf{0.991} & \textbf{0.825} \\
Swapped Inputs      & 0.745 & 0.952 & 0.982 & 0.725 \\
\hline
\end{tabular}
\end{table}

\subsection{Backbone Generalization}
\label{subsec:backbone}

To validate the backbone choice, we compared C-RADIO-B against DINOv2-B and DINOv3-B under identical conditions (frozen backbone, no SSL, \cref{tab:backbone}). C-RADIO-B achieves the strongest absolute correlation (0.816 with fusion) and the largest fusion gain (+0.022 over Detection-Only), suggesting its multi-teacher distillation (CLIP, DINOv2, SAM) produces features that are particularly amenable to the complementary fusion of global and localization signals. DINOv2-B remains competitive (0.806), while DINOv3-B shows a slight negative fusion delta ($-$0.003), indicating that not all foundation backbones benefit equally from our dual-pathway design.

\begin{table}[t]
\caption{Backbone comparison (all frozen, no SSL). C-RADIO-B achieves the best correlation and the largest fusion gain.}\label{tab:backbone}
\centering
\begin{tabular}{|l|c|c|c|}
\hline
\textbf{Backbone} & \textbf{Det-Only} & \textbf{Fusion} & $\Delta$\textbf{Corr.} \\
\hline
C-RADIO-B & \textbf{0.794} & \textbf{0.816} & +0.022 \\
DINOv2-B  & 0.791 & 0.806 & +0.015 \\
DINOv3-B  & 0.785 & 0.782 & $-$0.003 \\
\hline
\end{tabular}
\end{table}

%% ============================================================================
\section{Discussion and Limitations}
\label{sec:discussion}
%% ============================================================================

\textbf{Label definition.} Our quality labels use rule-guided heuristics grounded in clinical criteria rather than end-to-end radiologist preference. While inter-observer agreement is strong ($\kappa = 0.76$--$0.92$), rule-based scores simplify nuanced clinical judgments. Two factors mitigate this: (i)~Path~B relies on predicted landmarks, requiring genuine visual understanding; (ii)~strong generalization to unseen external data (\cref{subsec:external_images,subsec:video_validation}) suggests the model captures patterns beyond rule artifacts. Future work should incorporate direct expert preference labels and rank-based learning objectives.

\textbf{Why BYOL helps.}
BYOL's exponential moving average (EMA) target network stabilizes training even with small batch sizes (12 in our setup), avoiding the large-batch requirements of contrastive methods. Its positive-pair-only objective sidesteps hard-negative pitfalls that arise when visually similar frames differ subtly in clinical quality---a common scenario in fetal ultrasound where adjacent frames may look nearly identical yet vary in diagnostic suitability. In-domain SSL on 19K ultrasound images adapts backbone features to speckle texture, probe-specific contrast patterns, and anatomical priors absent from natural-image pretraining. The result is a +0.018 composite improvement over the no-SSL baseline (\cref{tab:ablation_arch}), with the largest per-task gains on the two hardest targets, NT and Scalebar (\cref{tab:pertask_auroc}).

\textbf{Clinical implications.}
FetSelect's real-time inference ($\sim$111\,FPS) and lightweight task heads ($\sim$2.2M parameters) make deployment feasible on clinical workstations or edge devices attached to ultrasound machines. In practice, the system could (i)~provide live frame-quality feedback during scanning, reducing the need for repeat sweeps; (ii)~automatically select optimal frames for downstream biometry pipelines, removing operator-dependent pre-selection; and (iii)~serve as a quality gate in retrospective studies, filtering suboptimal frames before measurement extraction. The task-specific design is critical: a global quality score would miss that a frame excellent for CRL may be poor for NT.

\textbf{Baseline comparisons.} No existing method addresses the same multi-task, task-specific frame selection problem, making direct comparison difficult. We provide adapted baselines (Random, Linear, Detection-Only) and comprehensive ablations (\cref{tab:ablation_arch,tab:pertask_auroc}). Broader comparison with adapted methods from adjacent domains (e.g., second-trimester plane classification, echocardiography quality scoring), and with CLIP-style zero-shot rankers built on FetalCLIP~\cite{maani2025fetalclip}, would further strengthen the evaluation and is a priority for future work.

\textbf{Statistical significance.} With $n{=}3$ seeds, $t$-tests have limited power. We report $p$-values and effect sizes (Cohen's $d = 0.6$--$1.1$); the consistent directionality across all four tasks supports practical significance. We acknowledge this limitation and recommend future work with $n \geq 5$ seeds or cross-validation.

\textbf{Other limitations.} Single primary annotator (validated by two senior experts); single-site acquisition; no temporal modeling across frames. The system currently treats each frame independently--incorporating temporal context (e.g., via sliding-window attention or recurrent modules over video clips) could improve robustness by leveraging motion continuity and probe trajectory.

%% ============================================================================
\section{Conclusion}
\label{sec:conclusion}
%% ============================================================================

We presented FetSelect, a multi-task architecture for automated best-frame selection in first-trimester fetal ultrasound. On a held-out test set, FetSelect achieves mean AUROC 0.956, Corr.\ 0.818 (composite 0.887), with temporally coherent behavior on external videos and robust task-aware generalization to unseen public images. Ablations demonstrate that both dual-pathway fusion and BYOL pretraining contribute complementary gains, with the largest improvements on the most clinically challenging targets (NT and Scalebar). The backbone comparison confirms that multi-teacher distilled features (C-RADIO-B) are particularly suited to our fusion design. Real-time inference ($>$100\,FPS) supports integration into clinical workflows. Future work will incorporate temporal modeling over video sequences, multi-site validation across scanner vendors, richer label sources including direct expert preference rankings, and extension to additional biometry targets beyond the first trimester. Code and pretrained weights are available at \url{https://github.com/mahmoodphd/FetSelect}.

\begin{credits}
\subsubsection{\ackname}
This publication was funded by the PPM7th Cycle grant
(PPM 07-0409-240041, AMAL-For-Qatar) from the Qatar Research Development
and Innovation Council (QRDI), a member of the Qatar Foundation.

\subsubsection{\discintname}
The authors have no competing interests to declare that are relevant to the content of this article.
\end{credits}

\bibliographystyle{splncs04}
\bibliography{mybib}

\end{document}